\documentclass[10pt]{article}

\usepackage[T1]{fontenc}
\usepackage[letterpaper,top=0.72in,bottom=0.78in,left=0.72in,right=0.72in]{geometry}
\usepackage[round,authoryear]{natbib}

\usepackage{amsmath,amssymb,booktabs,multirow}
\usepackage{graphicx}
\usepackage{microtype}
\usepackage{xcolor}
\usepackage{flafter}
\usepackage{float}
\usepackage{placeins}
\usepackage{xurl}
\usepackage[hidelinks]{hyperref}

\hypersetup{
  pdftitle={A Drosophila Whole-Connectome Network Can Learn Human-Designed Cognitive Tasks},
  pdfauthor={Joonghui Cho, Minchan Kang, and Daeshik Kim}
}

\newcommand{\pp}{\,pp}
\begin{document}

\twocolumn[
\begin{center}
{\LARGE\bfseries A Drosophila Whole-Connectome Network Can Learn\\[0.15em]
Human-Designed Cognitive Tasks\par}
\vspace{0.9em}
{\large Joonghui Cho \qquad Minchan Kang \qquad Daeshik Kim$^{*}$\par}
\vspace{0.25em}
{\normalsize Korea Advanced Institute of Science and Technology (KAIST)\par}
\vspace{0.15em}
{\small \href{mailto:paul26@kaist.ac.kr}{paul26@kaist.ac.kr}\qquad
\href{mailto:mc.kang@kaist.ac.kr}{mc.kang@kaist.ac.kr}\qquad
\href{mailto:daeshik@kaist.ac.kr}{daeshik@kaist.ac.kr}\par}
{\small $^{*}$Corresponding author.\par}
\vspace{0.9em}
\begin{minipage}{0.92\textwidth}

\begin{abstract}
Can a biological wiring diagram serve as a useful computational substrate beyond the behaviors for which it evolved? We use the publicly released MaleCNS v1.0 connectome, reconstructed from a single adult male \emph{Drosophila} specimen, as the fixed recurrent topology of an artificial network. We train separate models for bounded addition and for a controlled grounded relational language task built from a fixed 100-word lexicon. In both models, one scalar is learned per anatomical edge. The anatomical graph reaches $92.77\%$ mean accuracy on held-out addition, compared with $67.93\%$ for directed degree-preserving rewires. On the strict paired language endpoint, which matches original and order-reversed scenes to their corresponding descriptions, it reaches $61.59\%$ across four fixed interfaces, compared with $44.17\%$ for matched rewires. At the canonical interface, it ranks first in a fixed 21-graph comparison. On the matched 48-group intervention subset, shuffling task-defined sensory features reduces its score from $60.94\%$ to $19.27\%$. Together, these results show that higher-order MaleCNS wiring provides a reusable inductive bias for bounded addition and grounded relational language.
\end{abstract}
\end{minipage}
\end{center}
\vspace{0.8em}
]

\section{Introduction}

Connectomes are anatomical records, but they can also be read as candidate architectures. Synapse-resolution reconstructions now cover nearly the entire adult fly brain and the complete adult male central nervous system (CNS), revealing recurrent organizations shaped by an animal's behavior \citep{dorkenwald2024,berg2026}. We ask whether that organization remains useful when connection strengths are relearned for problems outside the animal's natural repertoire.

Most connectome-constrained models have pursued biological prediction in visual and sensorimotor systems, causal-effect estimation, or whole-brain activity fitting \citep{lappalainen2024,shiu2024,pospisil2024,wang2026}. Work on artificial tasks is smaller and has adapted fly wiring to image classification, chess, simulated locomotion, and robot navigation \citep{yu2026,jin2026,wangchen2026}. Evidence about the topology itself is mixed. Random structures performed similarly to biological ones in one comparison, and another study found that shared initialization and degree-preserving nulls removed much of an apparent advantage \citep{pircher2021,dhiman2026}. Task performance by itself does not show whether the biological arrangement of edges matters.

We use the MaleCNS adjacency as the recurrent mask of an artificial network, with one learned scalar on each retained edge. Addition and grounded relational language are trained as separate models. Addition requires exact answers for unseen operand pairs. The language model receives controlled English alongside task-defined sensory events and must distinguish descriptions of held-out channel pairs and reversed event sequences. Orthogonal token codes contain no pretrained lexical semantics, so the mapping must be acquired from the generated corpus and sensory stream.

\begin{figure*}[t]
\centering
\includegraphics[width=0.98\textwidth]{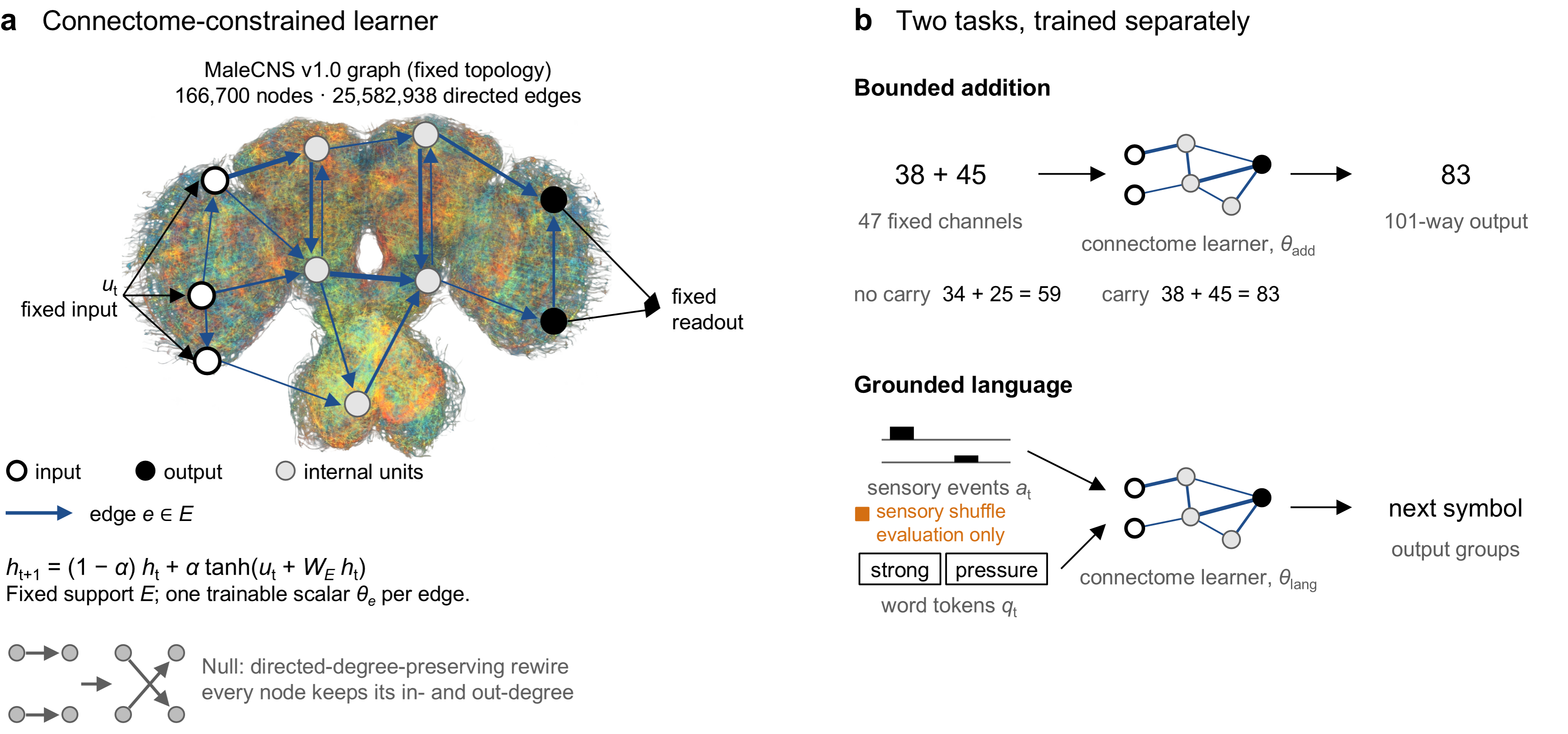}
\caption{Study design. (a) The processed MaleCNS v1.0 adjacency fixes the recurrent support, and each retained directed edge carries one learned scalar. The anatomical rendering shows a deterministic sample of 3,000 neuron skeletons. The overlaid nodes and arrows are schematic. Directed-degree-preserving rewires retain each node's in- and out-degree. (b) Addition and grounded relational language are separate fits on the same topology. Language receives sensory events and fixed token codes and predicts the next symbol through a fixed readout. Held-out evaluation pairs original and order-reversed scenes with their corresponding equal-length descriptions. Orange marks the sensory-shuffle control used only at evaluation.}
\label{fig:overview}
\end{figure*}

The anatomical graph outperforms degree-preserving rewires on both tasks, with the language separation appearing at every tested interface and in an additional 20-rewire ensemble. Sensory shuffling sharply reduces language performance. Figure~\ref{fig:overview} introduces the graph, tasks, and main controls.

\section{Related Work}

\paragraph{Whole-connectome resources and models.}
FlyWire reconstructed a whole adult female brain with 139,255 neurons and about 50 million chemical synapses \citep{dorkenwald2024}. MaleCNS extends synapse-resolution reconstruction across the complete adult male CNS \citep{berg2026}. Whole-brain analyses identify rich-club organization and overrepresented recurrent motifs \citep{lin2024}. Functional models use this anatomical detail to predict visual activity \citep{lappalainen2024}, model feeding and grooming transformations \citep{shiu2024}, and formulate estimators for causal neural effects \citep{pospisil2024}. BrainTrace, for example, applied its online learning method to a FlyWire-constrained spiking model of resting-state regional activity and interregional functional connectivity \citep{wang2026}.

\paragraph{Topology as an inductive bias.}
Connectome-constrained reservoir studies have examined time-series prediction and multifunctional dynamics across hyperparameter regimes \citep{morra2025}. In small fly-connectome subgraphs, raw task performance was generally comparable to random reservoirs, but weight-cost-normalized performance and robustness to neuron loss or parameter variation were higher \citep{mcallister2026}. Simplified whole-brain communication models likewise produced simulated region-selective activation patterns that were sensitive to degree-preserving rewiring \citep{zhang2026}. A recent fly-visual-system study found that advantages over a weak random control largely disappeared with shared from-scratch initialization and degree-preserving nulls \citep{dhiman2026}.

Artificial-task studies use a complete larval connectome as a reservoir for image classification and chess \citep{yu2026}, an adult-brain connectome as a learned controller for simulated fly locomotion \citep{jin2026}, and a fly-brain-derived recurrent network for robot navigation \citep{wangchen2026}. FlyGM also compares its locomotion controller with a degree-preserving rewire. The present experiments move away from embodied behavior to two symbolic mappings and ask whether their performance survives changes in token-input and output placement and graph draw.

\paragraph{Grounding, composition, and arithmetic.}
Fly-inspired language work has used a mathematical mushroom-body motif to learn sparse word representations from text \citep{liang2021}. Associations between words and referents can also be learned from paired egocentric visual and linguistic experience \citep{vong2024}, and pretrained instruction embeddings can support compositional generalization in sensorimotor networks \citep{riveland2024}. Benchmarks such as gSCAN, COGS, ReaSCAN, and Winoground test whether familiar components remain interpretable in new combinations or orders \citep{ruis2020,kim2020,wu2021,thrush2022}. Our language experiment uses orthogonal token codes, channel-pair holdouts, sensory interventions, and relation reversal for the same general reason. The addition experiment reports exact answers and carry-sensitive results, following work that separates predictive performance from mechanistic recovery \citep{nikankin2025}.

\section{Methods}

\subsection{Experimental design}

The two task models are trained independently on the anatomical topology and on directed degree-preserving rewires. Addition provides a repeated-run comparison across several graphs, whereas language uses a crossed design of five topologies and four input-output interfaces followed by a larger 21-graph comparison at interface I1. Rewiring changes the adjacency while retaining node and edge counts, every node's directed degree, the task interface, parameter count, initialization distribution, optimizer, and training budget.

\subsection{Connectome network and topology controls}

We use the public MaleCNS v1.0 release \citep{berg2026} and retain every downloaded neuron assigned a superclass. The processed graph has 166,700 nodes and 25,582,938 directed neuron-to-neuron edges, representing 124,177,617 synaptic contacts. Appendix~\ref{app:graph} gives the specimen, selection, orientation, and provenance details.

Let $G=(V,E)$ denote the directed graph. For a batch state $h_t\in\mathbb{R}^{|V|\times B}$, both task models use a leaky bounded recurrence
\begin{equation}
h_{t+1}=(1-\alpha)h_t+\alpha\tanh\!\left(u_t+W_Eh_t\right),
\label{eq:recurrence}
\end{equation}
where $\alpha$ is the leak coefficient, $u_t$ is the task-specific input drive, and $W_E$ has nonzeros only on $E$. For an edge $e=(j\!\to\! i)$, $(W_E)_{ij}=\theta_e/\sqrt{\max(d_i^{\mathrm{in}},1)}$, where $d_i^{\mathrm{in}}$ is the in-degree of node $i$. The recurrent gain is one in both tasks. Synaptic contact counts determine only the mask, and edge scalars are initialized independently. The edge vector $\theta\in\mathbb{R}^{|E|}$ is the only trainable tensor, with input and readout maps and codebooks held fixed.

Rewired controls are generated by directed double-edge swaps that preserve every node's unweighted in-degree and out-degree, retain the original self-loops, and introduce no duplicate edges. After five accepted swaps per original edge, only $0.588$--$0.594\%$ of source edges remain across the 20-graph ensemble. The comparison tests the collective contribution of higher-order properties, including reciprocity, motifs, paths, communities, and spectra, beyond directed degree. Appendix~\ref{app:null} gives the construction details and structural diagnostics.

\subsection{Bounded-addition protocol}

The task contains every ordered pair of nonnegative integers satisfying $a+b\leq100$. An expression $a+b$ is encoded in 47 fixed channels: two 22-dimensional digit codes and three operator channels. The network receives 64 recurrent updates, and logits for the 101 answers from 0 through 100 are averaged over the final four steps. Training minimizes cross-entropy plus a normalized ordinal CDF loss with unit coefficient. The 5,151 expressions are divided into 3,584 training, 775 validation, and 792 test examples. Commuted expressions are grouped before splitting, so no unordered operand pair occurs in more than one split. Each epoch visits all training expressions once without replacement, and the checkpoint with the lowest validation loss is selected before a single evaluation of the test set.

The comparison contains four optimization runs on the anatomical graph, four on the first rewire, and two each on two additional rewires, yielding an unbalanced graph-by-initialization design. We report the individual runs and their descriptive aggregate.

\subsection{Grounded relational language protocol}\label{sec:language-protocol}

The language task pairs sensory events with a question and a controlled English answer. In one held-out example, strong pressure on the right middle leg occurs before weak movement on the right front leg. Asked ``What did I feel?'', the target answer begins ``The strong pressure on the right middle leg happened before the weak movement on the right front leg.'' Table~\ref{tab:worked-example} gives the complete record and its order reversal.

Each event combines one of six leg locations with one of four signal kinds and has binary strength, duration, and speed attributes. The 24 location--kind channels drive disjoint groups of anatomically annotated MaleCNS leg afferents, comprising 2,837 neurons. Event attributes determine the amplitudes and timing of the sensory input. Answers are rendered from a 100-word lexicon using authored task conventions, while the model vocabulary also includes three punctuation marks and six control symbols. Two fixed $128\times109$ codebooks assign orthonormal input and output codes to these 109 symbols.

Let $a_t\in\mathbb{R}^{24}$ contain the task-defined sensory amplitudes and let $q_t$ be the one-hot indicator for the current symbol. With fixed sensory and token gains $g_s$ and $g_w$, the language drive is
\begin{equation}
u_t=g_sP_sa_t+g_wP_wE_{\mathrm{in}}q_t.
\label{eq:language-input}
\end{equation}
For sensory group $S_g$, $(P_s)_{ig}=|S_g|^{-1/2}$ when $i\in S_g$ and zero otherwise, while each code dimension is copied by $P_w$ to eight token-input nodes. Token nodes are sampled without replacement from nodes with outgoing edges after excluding sensory nodes, and eight output nodes per code dimension are sampled from nodes with incoming edges after excluding both earlier sets. The sensory, token-input, and output sets are therefore disjoint. The four interfaces reuse the sensory groups and codebooks but place token and output nodes differently.

If $O_d$ is the eight-node output group for code dimension $d$, the readout is
\begin{equation}
z_{t,d}=\frac{1}{8}\sum_{i\in O_d}h_{t,i},\qquad
\ell_t=\frac{E_{\mathrm{out}}^\top z_t}{\tau\max(\lVert z_t\rVert_2,10^{-6})}.
\label{eq:language-readout}
\end{equation}
Here $\ell_t$ contains the symbol logits and $\tau$ is the temperature.

The network first encodes the sensory sequence, then retains its recurrent state while reading the prompt and predicting the answer. During training, we supply a start symbol followed by the correct preceding answer symbols and minimize cross-entropy over answer symbols and the end symbol. Controlled text records instead provide next-symbol training without sensory input. Appendix~\ref{app:model} specifies the temporal schedule and loss normalization.

To test new combinations of familiar sensory channels, we partition the 276 unordered channel pairs into 228 training, 24 composition-validation, and 24 composition-test pairs before rendering any text. All orders and wordings of a pair inherit its partition. The deterministic generator contains 2 report families, 18 question-answer families, and 11 text families. It materializes 200,000 training records and 4,096 records for each of familiar-family validation, composition validation, and composition test. Appendix~\ref{app:model} gives the split rule and template families.

Each full run uses the same budget of four million processed tokens. The first quarter is a report-only warm-up, and the remainder follows a deterministic 60:20:20 token mixture of reports, question answering, and text. Primary evaluation uses the final checkpoint. The factorial crosses the four interfaces with the anatomical graph and four rewires under one shared initialization. Both this comparison and the 20-rewire ensemble reuse settings established during method development at the canonical interface I1.

\paragraph{Primary endpoint.}
For each of 192 held-out two-event groups, we pair the original scene and its order reversal with their corresponding equal-length descriptions while leaving the question unchanged. Scoring each scene with each candidate gives four teacher-forced sequence log probabilities, summed over target tokens. A group is correct only when both matched scores strictly exceed both mismatched scores. Thus the model must favor the appropriate description for each scene and the appropriate scene for each description. Ties are failures. The primary endpoint is the fraction of correct groups. Under four continuous exchangeable scores, its chance reference is $1/6=16.67\%$, calibrated by score-position permutations in Appendix~\ref{app:evaluation}.

\subsection{Controls and statistical analysis}

The sensory-shuffle intervention leaves prompts and candidates unchanged but replaces the complete task-defined sensory-amplitude tensor with that of the next evaluation group in a deterministic cycle. The intervention uses two groups from each of 24 held-out families, for 48 groups in total, and compares normal and shuffled evaluation on the same records. A parameter-matched dense recurrent control has width $\mathrm{round}(\sqrt{|E|})=5{,}058$ and approximately the same number of recurrent parameters, but far fewer recurrent units. Two additional fits use the anatomical graph at I1 and are trained and evaluated without the task-defined sensory channel.

For the pair-preserving counterfactual endpoint, let $r^{(c)}$ change one factor of record $r$, with rendered answers $y$ and $y^{(c)}$. Using mean teacher-forced log probability per target token, denoted $m$, a counterfactual pair is correct when $m(r,y)>m(r,y^{(c)})$ and $m(r^{(c)},y^{(c)})>m(r^{(c)},y)$. We average the resulting accuracies across strength, duration, speed, and order, each evaluated on 48 groups. For comparison, we also report the stricter four-inequality score obtained by adding the two cross-scene comparisons from the primary endpoint. Location and signal-kind changes may alter channel-pair membership and are therefore excluded from the held-out aggregate.

The language factorial estimand is the mean within-interface difference between the anatomical graph and four rewires. We report a 20,000-sample two-way crossed bootstrap that independently resamples the four interface levels and the four rewire identities, preserving the crossed design. With four levels on each axis, the interval summarizes variation within the executed design. A separate analysis at I1 ranks the anatomical graph against 20 independently generated directed degree-preserving rewires under a shared initialization. Writing $T(G)$ for a graph's strict paired score, its finite-ensemble upper-tail fraction is
\begin{equation}
\widehat r_{20}=\frac{1+\#\{T(G_b^*)\geq T(G)\}}{20+1}.
\label{eq:rankp}
\end{equation}
The minimum attainable rank fraction is $1/21$. Two additional initializations test optimization robustness for the anatomical graph and two rewires.

\section{Results}

Table~\ref{tab:summary} summarizes the primary topology comparisons, and the accompanying figures show the individual optimization runs and graph draws.

\begin{table}[t]
\caption{Primary topology comparisons.}\label{tab:summary}
\centering
\footnotesize
\begin{tabular}{lrrr}
\toprule
Endpoint & Anat. & Rewired & $\Delta$ \\
\midrule
Addition overall & 92.77 & 67.93 & +24.84 \\
Addition carry ($<100$) & 92.96 & 69.09 & +23.88 \\
Language factorial & 61.59 & 44.17 & +17.42 \\
Language ensemble & 71.35 & 46.56 & +24.79 \\
\bottomrule
\end{tabular}
\vspace{2pt}
\begin{minipage}{0.96\linewidth}\footnotesize
Values are percentages. Addition averages four anatomical fits and eight fits over three rewired graphs. The language factorial crosses four interfaces with four rewires; the ensemble row compares one anatomical run with 20 independent rewires. The last column is descriptive.
\end{minipage}
\end{table}

\subsection{The connectome topology improves held-out addition}

\begin{figure}[!t]
\centering
\includegraphics[width=\linewidth]{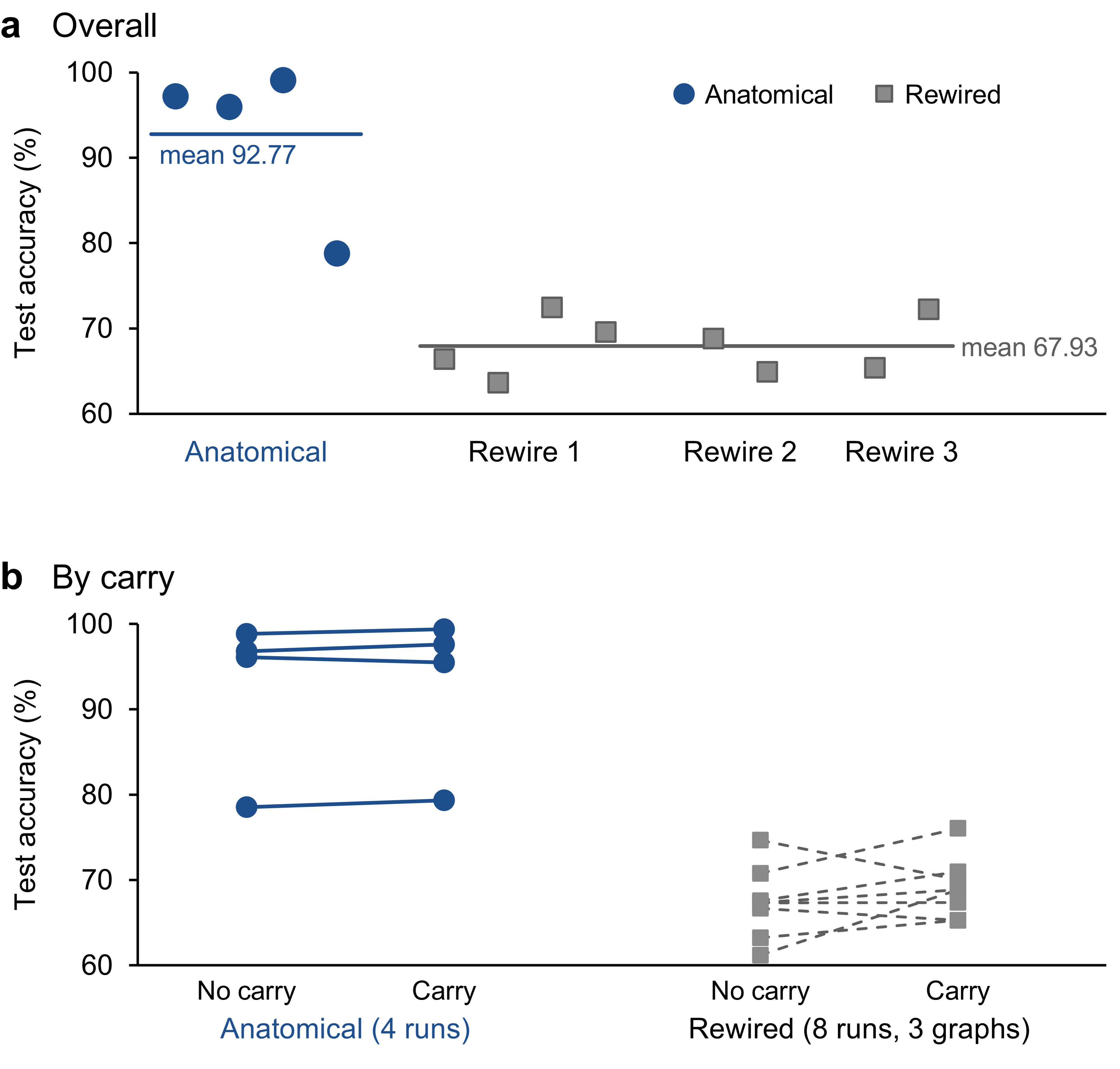}
\caption{Bounded-addition results. (a) Held-out accuracy for four anatomical runs and eight runs on three degree-preserving rewires. Horizontal segments mark topology-class means. (b) Accuracy without and with a ones-digit carry among examples whose sum is below 100. Lines connect the two strata within each run. Sum-100 examples are reported separately in Appendix~\ref{app:results}.}
\label{fig:addition}
\end{figure}

The anatomical graph reaches a mean held-out accuracy of $92.77\%$ across four optimization runs, compared with $67.93\%$ across eight fits on three rewired graphs, a difference of $24.84\pp$ (Table~\ref{tab:summary} and Figure~\ref{fig:addition}). On examples with a ones-digit carry and a sum below 100, the corresponding means are $92.96\%$ and $69.09\%$. The weakest anatomical run reaches $78.79\%$, indicating substantial optimization variability. The constant-answer test baseline is $0.51\%$. Appendix~\ref{app:results} reports every run, carry stratum, and mean absolute error.

\subsection{The topology advantage extends to grounded relational language}

\begin{figure*}[!t]
\centering
\includegraphics[width=\textwidth]{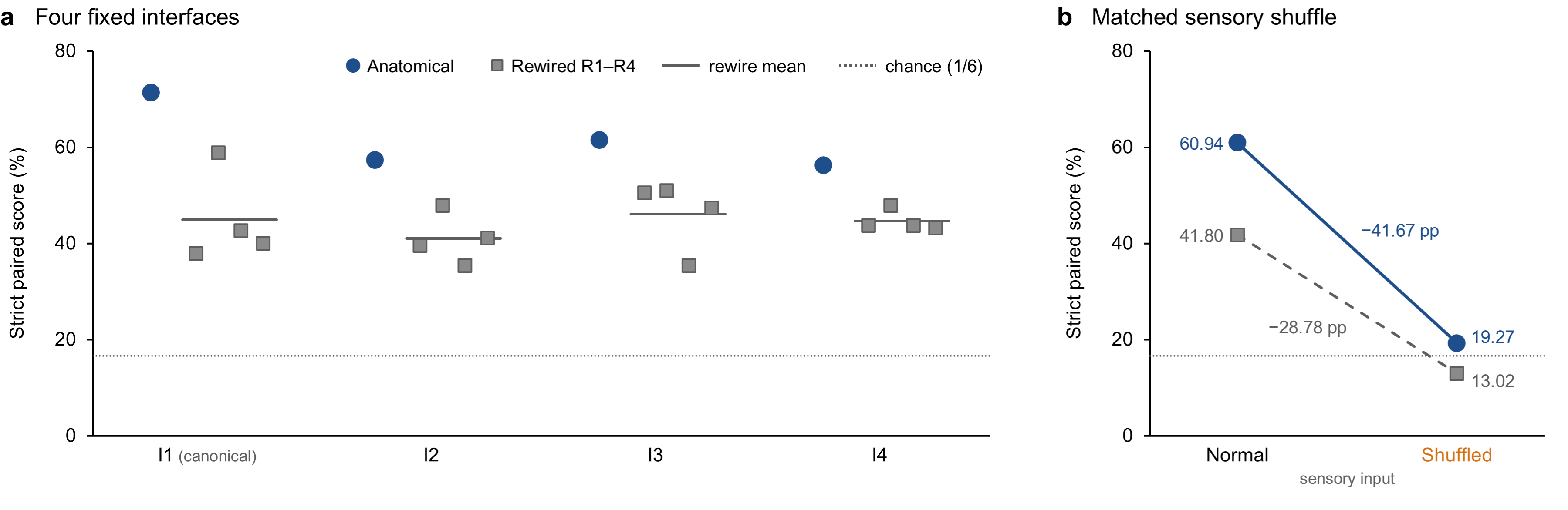}
\caption{Grounded relational language results. (a) Strict paired scores at the canonical interface I1 and three alternative placements. Each anatomical fit is compared with four rewires. Horizontal segments mark rewire means. (b) On the same 48 groups, sensory shuffling reduces the anatomical score from $60.94\%$ to $19.27\%$ and the rewire mean from $41.80\%$ to $13.02\%$. The dotted line marks the $1/6$ chance reference.}
\label{fig:language}
\end{figure*}

Varying the interface tests whether the topology difference persists across token-input and output placements. Across four interfaces, the anatomical graph obtains $61.59\%$ on the strict paired language endpoint. The four rewires paired with each interface average $44.17\%$, yielding a $17.42\pp$ difference and a 95\% crossed-bootstrap interval of $[11.39, 24.90]\pp$. The separation is positive at every placement tested (Figure~\ref{fig:language}).

\begin{figure}[!t]
\centering
\includegraphics[width=\linewidth]{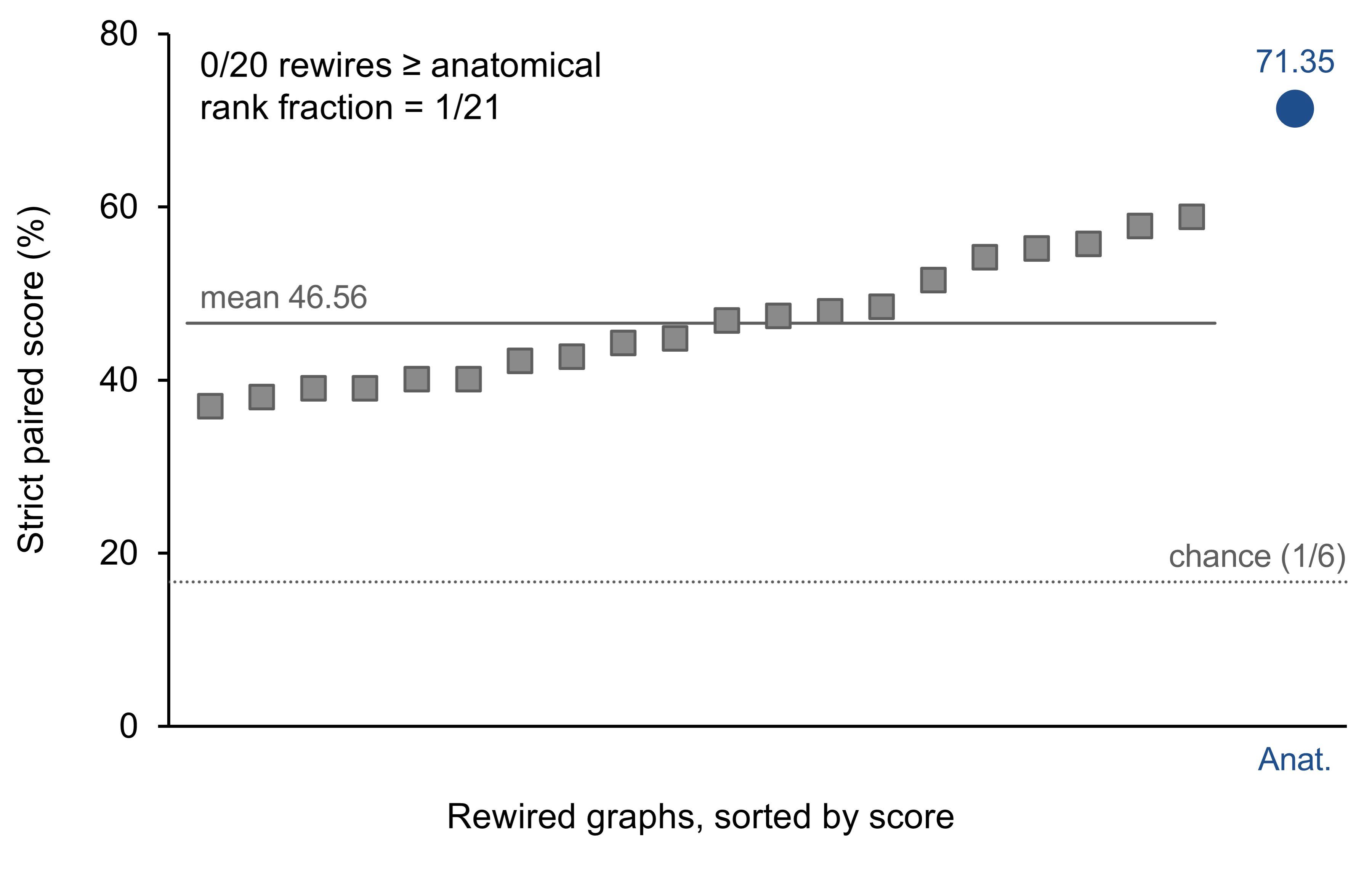}
\caption{Rewire ensemble at interface I1. The 20 degree-preserving rewire scores are sorted for display. The anatomical graph scores $71.35\%$, above every rewire and the ensemble mean of $46.56\%$. Its finite-ensemble rank fraction is $1/21$. The dotted line marks the $1/6$ chance reference.}
\label{fig:null}
\end{figure}

The larger ensemble examines variation across graph draws at I1, where the anatomical graph scores $71.35\%$. Twenty independent rewires average $46.56\%$ (SD $6.99\pp$, range $36.98$--$58.85\%$), and none reaches the anatomical score (Figure~\ref{fig:null}). The resulting finite-ensemble upper-tail rank fraction is $\widehat r_{20}=1/21=0.0476$.

Across three optimization runs at I1, the anatomical graph averages $71.01\%$ (SD $2.62\pp$), compared with $46.53\%$ for Rewire 1 (SD $8.87\pp$) and $46.18\%$ for Rewire 2 (SD $11.22\pp$). Optimization variance is smaller for the anatomical language runs than for either rewire, although the anatomical addition runs show a wider spread.

We next examined whether differences in access to the output nodes could explain the language gap. Every output node is reachable from both token and sensory inputs at all four interfaces. The largest strongly connected component contains $99.17\%$ of nodes in both topology classes, and the mean minimum token-to-output distance is $1.51$ in both. The anatomical graph has far more reciprocal directed edges ($29.89\%$ versus $0.514$--$0.520\%$) and longer minimum sensory-to-output paths (mean $2.55$ versus $1.38$). Table~\ref{tab:structural} gives the full structural comparison.

\subsection{Sensory alignment and relational generalization}

On the 48-group intervention subset, shuffling the task-defined sensory amplitudes reduces the anatomical score from $60.94\%$ to $19.27\%$, a $41.67\pp$ drop. The mean rewired score falls from $41.80\%$ to $13.02\%$, a $28.78\pp$ drop. Two models trained without sensory input score $0\%$: original and reversed scenes then produce identical inputs, making the endpoint's four strict inequalities impossible to satisfy. The shuffle result connects language performance to the alignment between the sensory stream and the rendered descriptions.

The counterfactual evaluation asks whether changing one scene factor changes which answer the model prefers. On the two-preference pair-preserving endpoint for strength, duration, speed, and order, the anatomical graph reaches $34.11\%$ versus $8.66\%$ for rewires. The corresponding strict four-comparison means are $16.41\%$ and $4.69\%$. Anatomical accuracy on the two-preference endpoint is highest for duration and strength and lowest for speed (Table~\ref{tab:counterfactual-factors}). For comparison, the parameter-matched dense recurrent model obtains $8.72\%$ on the primary endpoint, $1.56\%$ under sensory shuffle, and $3.39\%$ on the two-preference counterfactual endpoint.

\section{Discussion}

Across the two tasks, the anatomical graph's advantage indicates that directed degree does not capture all of the useful organization in this connectome. The language gap persists across token-input and output placements and graph draws while sensory afferents remain in their anatomical groups. Because each task is fitted independently, what carries across tasks is the recurrent architecture rather than trained weights.

The language results also depend on the correspondence between sensory input and text. The attribute-level counterfactual scores show that this grounding is uneven (Table~\ref{tab:counterfactual-factors}). The anatomical model distinguishes strength and duration changes more reliably than changes in speed.

The structural diagnostics make a simple reachability explanation unlikely, since both topology classes retain access from the inputs to every output node. Rewiring changes the organization of those routes even when access is preserved. Null models that retain reciprocity, community structure, or sensory-to-output path statistics could distinguish their contributions to the performance gap.

\section{Limitations}

We evaluate one anatomical graph and two synthetic tasks. The addition runs form an unbalanced graph-by-initialization design, and the 20-graph language ensemble limits rank resolution to $1/21$. Four interfaces sample only a small part of the possible placement space.

The compact vocabulary, authored semantics, and controlled grammar make relational generalization measurable but leave open scaling to natural language. The network omits synaptic signs, neurotransmitter identities, delays, cell dynamics, and biological plasticity. Its dense control matches recurrent parameter count rather than state dimension or sparsity.

\section{Conclusion}

A recurrent network with the processed MaleCNS topology can learn bounded addition and controlled grounded relational language through separate fits. Degree-preserving rewiring reduces performance on both tasks, with the language gap persisting across the tested token-input and output placements and graph draws. Sensory shuffling sharply reduces language performance, linking the learned descriptions to the task-defined input stream. These results identify the connectome as a useful architectural prior and motivate isolating the structural properties responsible for its advantage.

\FloatBarrier
\section*{AI Use Statement}

We used OpenAI generative AI tools to develop and review the synthetic data generator and to polish the manuscript's prose. Their use for the training code, equations, literature, methodology, result interpretation, figures, and translations was limited to review. The authors take responsibility for the final content.

\clearpage
\appendix
\onecolumn
\begin{center}
{\Large\bfseries Supplementary Materials\par}
\vspace{0.45em}
{\large A Drosophila Whole-Connectome Network Can Learn Human-Designed Cognitive Tasks\par}
\end{center}
\vspace{0.8em}

\section{Data and Graph Provenance}\label{app:graph}

MaleCNS v1.0 reconstructs the CNS of a single adult male and was obtained from the official release page \citep{berg2026}. The MaleCNS v1.0 dataset is available under CC BY 4.0.

We selected annotation rows with an assigned superclass, sorted retained body IDs, and kept a directed connection only when both endpoints survived selection. Repeated records for the same ordered neuron pair were aggregated by summing their contact counts. Contact counts define edge presence rather than initial weight magnitudes. Our rule retains 166,700 unique body IDs, matching the total reported by \citet{berg2026}. Their connectivity graph contains 166,483 neurons after excluding 217 neurons without synaptic connections. Table~\ref{tab:graph} summarizes the source and processed graph counts.

\begin{table}[h]
\caption{Processed graph and source statistics.}\label{tab:graph}
\centering
\begin{tabular}{lr}
\toprule
Quantity & Value \\
\midrule
Source-article neuron total & 166,700 neurons \\
Source-article connected neurons & 166,483 neurons \\
Selection & Superclass annotation assigned \\
Selected neurons in processed graph & 166,700 \\
Directed aggregated edges & 25,582,938 \\
Synaptic contacts represented & 124,177,617 \\
Self-edges & 101 \\
Edge threshold & None \\
Orientation & postsynaptic rows, presynaptic columns \\
\bottomrule
\end{tabular}
\end{table}

Graph orientation follows the implemented sparse matrix: rows are postsynaptic neurons and columns are presynaptic neurons. Thus a stored edge $(i,j)$ contributes from neuron $j$ to neuron $i$. The graph contains 101 self-edges, which remain unchanged under rewiring. Appendix~\ref{app:null} gives the rewire algorithm and structural diagnostics.

The anatomical rendering in Figure~\ref{fig:overview} uses a deterministic sample of 3,000 retained neurons with nonnull soma locations from the official precomputed skeleton service. The overlaid nodes and arrows illustrate the network schematically.

\section{Task, Model, and Training Details}\label{app:model}

\subsection{Shared recurrent parameterization}

In both tasks, the edge scalars are initialized independently from $\mathcal{N}(0,1)$ using a run-specific random state. The factor $1/\sqrt{\max(d_i^{\mathrm{in}},1)}$ normalizes every edge entering node $i$, and recurrent states begin at zero. Input groups, output groups, graph indices and language codebooks are fixed, so only the edge scalars are optimized.

\subsection{Addition task and optimization}

Input groups contain 64 graph nodes per channel, and output groups contain 32 nodes per class. The groups are sampled without replacement from nodes with outgoing or incoming edges and are disjoint. For an encoded expression $x\in\mathbb{R}^{47}$, channel $x_c$ is copied with amplitude 1 to every node in its input group at each recurrent update. If $R_k$ is the output group for answer $k$, its logit is
\[
\ell_k=\frac{1}{4\tau}\sum_{t=T-3}^{T}\frac{1}{|R_k|}\sum_{i\in R_k}h_{t,i},
\qquad T=64,
\]
with temperature $\tau=0.1$. The readout has no learned weight or bias.

We use $\alpha=0.5$, unit recurrent gain, and AdamW with learning rate $10^{-3}$ and weight decay $10^{-4}$. The gradient norm is clipped at 1, and the microbatch size is 128. Runs train for at most 100 epochs, with a minimum of 20 epochs and patience 15. If $p=\operatorname{softmax}(\ell)$ and $y\in\{0,\ldots,100\}$, the training loss is
\[
\mathcal{L}_{\mathrm{add}}
=-\log p_y+\frac{1}{100}\sum_{c=0}^{99}
\left(\sum_{k=0}^{c}p_k-\mathbf{1}\{y\leq c\}\right)^2.
\]
The second term penalizes disagreement between the predicted and target cumulative distributions over the ordered answer classes.

\subsection{Language task specification}

An event is a tuple $e=(l,k,v,d,s)$. Here $l$ is one of six leg locations, $k$ is one of four signal kinds, and $v$, $d$, and $s$ are binary strength, duration, and speed attributes. The channel index is $4l+k$, giving 24 sensory channels. Complete report answers describe all five attributes. Table~\ref{tab:worked-example} gives a held-out composition-test record. The primary paired endpoint renders both descriptions in a fixed relation-first form (``$X$ happened before $Y$'') and leaves the question unchanged after reversing event order.

\begin{table}[ht]
\caption{Worked example of a paired language group.}\label{tab:worked-example}
\centering
\footnotesize
\begin{tabular}{p{.17\textwidth}p{.77\textwidth}}
\toprule
Component & Content \\
\midrule
Question & What did I feel? \\
Original scene $r$ & strong, long, slow pressure on the right middle leg; then weak, short, slow movement on the right front leg \\
Candidate $y$ & The strong pressure on the right middle leg happened before the weak movement on the right front leg. The first signal started slowly and stopped after a long time. The last signal started slowly and stopped after a short time. \\
Reversed scene $r^{\leftrightarrow}$ & weak, short, slow movement on the right front leg; then strong, long, slow pressure on the right middle leg \\
Candidate $y^{\leftrightarrow}$ & The weak movement on the right front leg happened before the strong pressure on the right middle leg. The first signal started slowly and stopped after a short time. The last signal started slowly and stopped after a long time. \\
\bottomrule
\end{tabular}
\end{table}

Table~\ref{tab:authored-semantics} lists the additional meaning and action conventions used by a subset of question-answer and text families.

\begin{table}[ht]
\caption{Authored signal conventions in the language generator.}\label{tab:authored-semantics}
\centering
\small
\begin{tabular}{lll}
\toprule
Signal kind & Rendered meaning & Rendered action \\
\midrule
touch & a signal from the body & follow the signal \\
taste & food & approach the food \\
movement & a change in the ground & follow the movement \\
pressure & pain and danger & avoid that place \\
\bottomrule
\end{tabular}
\end{table}

The complete lexical vocabulary is shown in Table~\ref{tab:lexicon}. Punctuation and the six control symbols are separate model symbols and are not counted among the 100 words. The input and output codebooks have orthonormal columns and are generated once for use at all four interface placements. Sensory waveforms are deterministic functions of the split index and record identifier.

\begin{table}[ht]
\caption{The 100-word lexicon.}\label{tab:lexicon}
\centering
\footnotesize
\begin{tabular}{p{.14\textwidth}p{.80\textwidth}}
\toprule
Category & Words \\
\midrule
Nouns & fly, body, leg, legs, side, strength, touch, taste, movement, pressure, signal, sensation, event, events, pain, food, ground, danger, place, time, pulse, change \\
Attributes & left, right, front, middle, hind, weak, strong, short, long, quickly, slowly, first, last, same, different, one, two, three, four \\
Verbs & feel, felt, detect, detected, happen, happened, start, started, stop, stopped, mean, meant, avoid, avoided, approach, approached, follow, followed, do \\
Function words & a, the, that, my, its, it, i, they, both, another, is, are, was, were, did, in, on, to, from, of, and, but, so, because, before, after, while, then, not, no, yes, what, where, when, which, why, how, many, there, or \\
\bottomrule
\end{tabular}
\end{table}

The generator contains the families summarized in Table~\ref{tab:families}. Report families express complete event descriptions and supply the composition endpoint. Question-answer families isolate attributes, comparisons, temporal order, meanings, and actions. Text families provide next-symbol training on short controlled documents.

\begin{table}[ht]
\caption{Template families used by the deterministic corpus generator.}\label{tab:families}
\centering
\footnotesize
\begin{tabular}{p{.16\textwidth}p{.06\textwidth}p{.66\textwidth}}
\toprule
Mode & Count & Content \\
\midrule
Report & 2 & First- and third-person descriptions of complete event sequences \\
Question answering & 18 & Event identity and attributes, temporal order, location, comparisons, counts, meaning, action, and causal explanation \\
Controlled text & 11 & Temporal and causal statements, contrasts, simultaneous events, counting, side descriptions, pulses, and short dialogue \\
\bottomrule
\end{tabular}
\end{table}

\FloatBarrier
\subsection{Corpus generation and split construction}

The generator indexes the 24 sensory channels from 0 to 23. It assigns the 24 adjacent cyclic pairs $\{i,(i+1)\bmod24\}$ to composition validation and the 24 offset-five pairs $\{i,(i+5)\bmod24\}$ to composition test after canonicalizing pair order, leaving the remaining 228 unordered pairs for training. Scenes are sampled before wording is rendered, so every event order and paraphrase inherits the same pair partition. Familiar-family validation uses a deterministic hash reservation over attributed scenes and additionally reserves prompt identities for text records.

Training records consist of reports ($60\%$), question answering ($20\%$), and controlled text ($20\%$). Training reports contain one or two events, whereas composition-test records contain two. Selected text families contain three or four events. Two-event reports and the two report-based question-answer families use four surface variants, while one-event reports use a single form. Other question-answer and text families use a single template, except dialogue records that combine report variants with question families. The materialized bank contains 200,000 training records and 119,040 distinct conditional examples after exact deduplication. The first quarter of the token budget draws only from reports. The remainder selects the currently under-exposed pool to maintain the same 60:20:20 ratio by processed tokens.

\subsection{Language interface and optimization}

The sensory interface contains 24 disjoint anatomical afferent groups comprising 2,837 neurons selected using the side, leg entry nerve, class and subclass annotations. The task channels correspond to tactile mechanosensory bristles (touch), gustatory leg bristles (taste), chordotonal organs (movement) and campaniform sensilla (pressure). Each group is scaled by the inverse square root of its size. Token codes have dimension 128, with eight token-input nodes and eight output nodes per dimension. Token-input and output groups follow the sampling procedure described in Section~\ref{sec:language-protocol}.

Each sensory event occupies eight encoding steps. Its active channel lasts two steps for a short event or four for a long event. Weak and strong events have base amplitudes 0.4 and 0.8. Slow events use a linear envelope from 0.6 to 1, whereas fast events use a flat envelope. Deterministic Gaussian noise with standard deviation 0.003 is added and clipped to $[0,1]$. Events are encoded in scene order.

After sensory encoding, the grounded-mode symbol is applied for one recurrent transition. The network then processes the prompt followed by a separator and predicts the answer from a start symbol. Each prompt symbol and teacher-forced previous-target symbol receives four recurrent transitions with a one-step input pulse. Text records use the text-mode symbol and predict the complete document from the start symbol without a conditioning prompt or sensory input. Five non-output control symbols are masked from the logits, while punctuation, lexical words, and the end symbol remain available. The recurrence uses leak 0.5, token gain 4, sensory gain 1, and temperature 0.1.

AdamW uses learning rate $10^{-3}$, weight decay $10^{-4}$, global batch 192, microbatch 24, and gradient clipping at 1. Microbatch gradients are accumulated with summed cross-entropy normalized by the total number of supervised symbols in the global batch, excluding padding. Primary results use the final checkpoint after the common four-million-token training budget.

\subsection{Hyperparameter selection and interfaces}

Method development established the language recurrence, optimizer, curriculum, codebook construction, and canonical interface I1. A training-only diagnostic checked task learnability. The subsequent four-interface factorial and 20-rewire ensemble reused these settings and the same evaluation protocol and training budget.

\subsection{Language controls}

The dense control replaces the sparse recurrence with a fully connected 5,058-unit state and a $5{,}058\times5{,}058$ trainable recurrent matrix, initialized from $\mathcal{N}(0,1)$ and scaled by $1/\sqrt{5{,}058}$. It shares the codebooks, task generator, encoding schedule, objective, optimizer settings, curriculum, and four-million-token budget of the connectome models. Its sensory interface consists of 24 singleton input groups. Token and output groups retain eight units per code dimension. Two no-sensory controls use the anatomical graph at I1 and were trained from scratch with sensory input disabled. Evaluation also holds the sensory tensor at zero.

\FloatBarrier

\section{Evaluation Metrics and Statistical Procedures}\label{app:evaluation}

\subsection{Strict paired language endpoint}

For a held-out two-event record $r$, let $r^{\leftrightarrow}$ reverse the event order without changing the question. Let $y$ and $y^{\leftrightarrow}$ be the equal-length answers rendered from the two scenes. If $s(x,y)$ is the teacher-forced sequence log probability of candidate $y$ under scene and prompt $x$, summed over target symbols, the group is correct exactly when
\begin{align*}
s(r,y)&>s(r,y^{\leftrightarrow}), & s(r^{\leftrightarrow},y^{\leftrightarrow})&>s(r^{\leftrightarrow},y),\\
s(r,y)&>s(r^{\leftrightarrow},y), & s(r^{\leftrightarrow},y^{\leftrightarrow})&>s(r,y^{\leftrightarrow}).
\end{align*}
Thus both matched scores must exceed both mismatched scores, and ties fail. The primary score is the proportion correct over 192 held-out groups. With four continuous exchangeable scores, four of the $4!=24$ orderings satisfy this criterion, yielding the analytic reference $1/6$. Candidate lengths are equal within each group, so summed and mean log probabilities give the same ordering for this endpoint.

The sensory intervention uses two records from each of the 24 held-out pair families. For each of these 48 group IDs, the normal evaluation is compared with a shuffle that substitutes the full sensory-amplitude tensor from the next group in a deterministic cycle. The prompt and candidates do not change, so both scores are computed on the same records.

For a pair-preserving counterfactual $r^{(c)}$, the two-preference criterion requires
\[
m(r,y)>m(r,y^{(c)})\quad\text{and}\quad
m(r^{(c)},y^{(c)})>m(r^{(c)},y),
\]
where $m$ is mean teacher-forced log probability per target symbol. The strict version adds the two cross-scene inequalities used above. Strength, duration, speed, and order each use 48 groups. Location and signal-kind substitutions are excluded because they may move a record to a different held-out channel-pair family.

\subsection{Factorial uncertainty and graph ranking}

Let $D_{ij}$ be the score difference between the anatomical graph and rewire $j$ at interface $i$. The reported factorial effect is $\bar D=16^{-1}\sum_{i=1}^4\sum_{j=1}^4D_{ij}$. For each of 20,000 bootstrap replicates, four interfaces and four rewires are independently resampled with replacement, and the mean of the resulting crossed $4\times4$ table is recorded. Its 2.5th and 97.5th percentiles give the reported interval for variation within this $4\times4$ design.

The I1 ensemble compares one anatomical graph with 20 independently constructed rewires. Its upper-tail rank fraction is
\[
\widehat r_{20}=\frac{1+\#\{T(G_b^*)\geq T(G)\}}{21}.
\]
The smallest attainable value is $1/21$, and we report the quantity as a finite-ensemble rank without assuming uniformity of the double-edge-swap chain. Endpoint calibration separately permutes the four score positions within each group 2,000 times, recovering the analytic $16.67\%$ reference at all four anatomical interfaces.

\section{Rewire Construction and Structural Diagnostics}\label{app:null}

Each control begins from the processed directed graph. A proposed double-edge swap replaces $(a,b)$ and $(c,d)$ with $(a,d)$ and $(c,b)$. Proposals that introduce a duplicate edge or change a self-edge are rejected. All other accepted swaps preserve every node's in-degree and out-degree. Each rewire uses five accepted swaps per source edge. The acceptance rate is approximately $98.66\%$ across the 20 controls.

We computed Table~\ref{tab:structural} directly from the saved edge arrays. Reciprocity is the fraction of directed edges whose reverse is present. Strongly connected components follow the presynaptic-to-postsynaptic direction. For each interface, the path statistic averages across output nodes the minimum unweighted directed distance from any node in the named input set. The sensory set contains 2,837 anatomical afferents; the token set contains 1,024 input nodes. Anatomical values average four interfaces. Rewire path values pool four interfaces and 20 graphs. Every output node was reachable.

\begin{table}[htbp]
\caption{Structural diagnostics for the anatomical graph and rewired controls. Brackets give the minimum and maximum across 20 rewires.}\label{tab:structural}
\centering
\small
\begin{tabular}{lrr}
\toprule
Diagnostic & Anatomical & 20 rewires \\
\midrule
Source-edge overlap (\%) & 100 & 0.591 [0.588, 0.594] \\
Accepted swaps per edge & 0 & 5 \\
Reciprocal directed edges (\%) & 29.893 & 0.517 [0.514, 0.520] \\
Largest SCC (\% nodes) & 99.169 & 99.176 [99.175, 99.176] \\
Token$\to$output min. distance & 1.507 & 1.506 [1.448, 1.546] \\
Sensory$\to$output min. distance & 2.552 & 1.378 [1.343, 1.412] \\
Output reachability (\%) & 100 & 100 \\
\bottomrule
\end{tabular}
\end{table}

Rewiring replaces almost every edge but leaves the giant strongly connected core and token-to-output distance nearly unchanged. It greatly reduces reciprocity and shortens sensory-to-output distances. Shared reachability but divergent higher-order statistics motivate more selective null models.

\section{Compute and Reproducibility}\label{app:compute}

Both tasks were implemented in PyTorch and trained in FP32, with each process using one NVIDIA RTX 3090 GPU. Sparse recurrent propagation costs $O(|E|B)$ per step and uses $O(|E|+|V|B)$ principal storage aside from autograd state.

\paragraph{Code and data availability.}
MaleCNS v1.0 source tables and skeletons are available from the \href{https://male-cns.janelia.org/download/}{official release page}. Code is available at \url{https://github.com/joonghui0926/drosophila-connectome-cognitive-tasks}. The repository includes compact result records, checksums and instructions for rebuilding the graph and generated corpus. MaleCNS data and full training checkpoints are not redistributed.

\section{Detailed Numerical Results and Controls}\label{app:results}

Unless otherwise noted, accuracy and score entries in Tables~\ref{tab:addall}--\ref{tab:dense-runs} are percentages. MAE is measured in answer-value units, and SD is reported in percentage points.

\subsection{Addition runs}

Table~\ref{tab:addall} reports every completed addition run used in the paper. The no-carry and carry columns partition the 772 expressions whose sum is below 100; the 20 expressions that equal 100 form a separate stratum. Means weight completed optimization runs equally.

\begin{table}[H]
\caption{Every bounded-addition test run.}\label{tab:addall}
\centering
\small
\begin{tabular}{llrrrrr}
\toprule
Topology & Run & Overall & No carry & Carry & Sum 100 & MAE \\
 & & (792) & (438) & (334) & (20) & \\
\midrule
Anatomical & 1 & 97.22 & 96.80 & 97.60 & 100 & 0.078 \\
Anatomical & 2 & 95.96 & 96.12 & 95.51 & 100 & 0.110 \\
Anatomical & 3 & 99.12 & 98.86 & 99.40 & 100 & 0.040 \\
Anatomical & 4 & 78.79 & 78.54 & 79.34 & 75 & 0.266 \\
Rewire 1 & 1 & 66.41 & 67.35 & 67.37 & 30 & 0.413 \\
Rewire 1 & 2 & 63.64 & 63.24 & 65.27 & 45 & 0.448 \\
Rewire 1 & 3 & 72.47 & 74.66 & 70.06 & 65 & 0.355 \\
Rewire 1 & 4 & 69.57 & 67.58 & 70.96 & 90 & 0.393 \\
Rewire 2 & 1 & 68.81 & 67.35 & 68.86 & 100 & 0.385 \\
Rewire 2 & 2 & 64.90 & 66.67 & 65.27 & 20 & 0.434 \\
Rewire 3 & 1 & 65.40 & 61.19 & 68.86 & 100 & 0.443 \\
Rewire 3 & 2 & 72.22 & 70.78 & 76.05 & 40 & 0.359 \\
\midrule
Anatomical mean & -- & 92.77 & 92.58 & 92.96 & 93.75 & 0.124 \\
Rewire mean & -- & 67.93 & 67.35 & 69.09 & 61.25 & 0.404 \\
\bottomrule
\end{tabular}
\end{table}

\FloatBarrier

\subsection{Language factorial and controls}

Table~\ref{tab:interfaces} reports every cell in the crossed topology-interface comparison. Table~\ref{tab:nullscores} separately lists the 20 rewire scores used for the graph-level rank statistic at I1.

\begin{table}[H]
\caption{Every cell in the primary language factorial. R1--R4 denote the four prespecified rewired graphs.}\label{tab:interfaces}
\centering
\small
\begin{tabular}{lrrrrrrr}
\toprule
Interface & Anatomical & R1 & R2 & R3 & R4 & Rewire mean & $\Delta$ \\
\midrule
I1 (canonical) & 71.35 & 38.02 & 58.85 & 42.71 & 40.10 & 44.92 & 26.43 \\
I2 & 57.29 & 39.58 & 47.92 & 35.42 & 41.15 & 41.02 & 16.28 \\
I3 & 61.46 & 50.52 & 51.04 & 35.42 & 47.40 & 46.09 & 15.36 \\
I4 & 56.25 & 43.75 & 47.92 & 43.75 & 43.23 & 44.66 & 11.59 \\
\midrule
Mean & 61.59 & 42.97 & 51.43 & 39.32 & 42.97 & 44.17 & 17.42 \\
\bottomrule
\end{tabular}
\end{table}

\begin{table}[H]
\caption{I1 rewire scores, ordered from lowest to highest.}\label{tab:nullscores}
\centering
\small
\begin{tabular}{rr@{\qquad}rr@{\qquad}rr}
\toprule
Position & Score & Position & Score & Position & Score \\
\midrule
1 & 36.98 & 8  & 42.71 & 15 & 51.56 \\
2 & 38.02 & 9  & 44.27 & 16 & 54.17 \\
3 & 39.06 & 10 & 44.79 & 17 & 55.21 \\
4 & 39.06 & 11 & 46.88 & 18 & 55.73 \\
5 & 40.10 & 12 & 47.40 & 19 & 57.81 \\
6 & 40.10 & 13 & 47.92 & 20 & 58.85 \\
7 & 42.19 & 14 & 48.44 &    &       \\
\bottomrule
\end{tabular}
\end{table}

\begin{table}[H]
\caption{Strict paired scores at I1 across optimization runs.}\label{tab:language-runs}
\centering
\small
\begin{tabular}{lrrrrr}
\toprule
Topology & Run 1 & Run 2 & Run 3 & Mean & SD \\
\midrule
Anatomical & 71.35 & 68.23 & 73.44 & 71.01 & 2.62 \\
Rewire 1 & 38.02 & 55.73 & 45.83 & 46.53 & 8.87 \\
Rewire 2 & 58.85 & 37.50 & 42.19 & 46.18 & 11.22 \\
\bottomrule
\end{tabular}
\end{table}

\begin{table}[H]
\caption{Sensory intervention by interface. Normal and shuffled values use the same records; rewire entries average four graphs.}\label{tab:sensory-by-interface}
\centering
\small
\begin{tabular}{lrrrr}
\toprule
& \multicolumn{2}{c}{Anatomical} & \multicolumn{2}{c}{Rewire mean} \\
Interface & Normal & Shuffled & Normal & Shuffled \\
\midrule
I1 & 68.75 & 25.00 & 44.79 & 12.50 \\
I2 & 56.25 & 18.75 & 39.58 & 13.02 \\
I3 & 62.50 & 20.83 & 42.71 & 14.58 \\
I4 & 56.25 & 12.50 & 40.10 & 11.98 \\
\midrule
Mean & 60.94 & 19.27 & 41.80 & 13.02 \\
\bottomrule
\end{tabular}
\end{table}

\noindent\begin{minipage}{\textwidth}
\begin{table}[H]
\caption{Pair-preserving counterfactual accuracy by changed factor. Each factor uses 48 groups per run.}\label{tab:counterfactual-factors}
\centering
\small
\begin{tabular}{lrrrr}
\toprule
& \multicolumn{2}{c}{Two-preference} & \multicolumn{2}{c}{Strict four-inequality} \\
Factor & Anatomical & Rewired & Anatomical & Rewired \\
\midrule
Strength & 46.35 & 13.02 & 17.19 & 1.82 \\
Duration & 59.38 & 4.69 & 22.40 & 0.52 \\
Speed & 4.17 & 0.52 & 0.00 & 0.13 \\
Order & 26.56 & 16.41 & 26.04 & 16.28 \\
\midrule
Macro mean & 34.11 & 8.66 & 16.41 & 4.69 \\
\bottomrule
\end{tabular}
\end{table}

The counterfactual result is concentrated in strength and duration. Speed remains near zero.
\end{minipage}

\begin{table}[H]
\caption{Dense-control results by optimization run. Primary, Normal, and Shuffled report strict paired accuracy. Counterfactual reports the two-preference macro accuracy over strength, duration, speed, and order. All values are percentages.}\label{tab:dense-runs}
\centering
\small
\begin{tabular}{lrrrr}
\toprule
Run & Primary & Normal (48) & Shuffled (48) & Counterfactual \\
\midrule
1 & 5.73 & 4.17 & 0.00 & 2.08 \\
2 & 8.33 & 6.25 & 0.00 & 3.65 \\
3 & 17.71 & 22.92 & 6.25 & 6.25 \\
4 & 3.13 & 4.17 & 0.00 & 1.56 \\
\midrule
Mean & 8.72 & 9.38 & 1.56 & 3.39 \\
\bottomrule
\end{tabular}
\end{table}

Both no-sensory checkpoints score $0\%$ on the primary and counterfactual endpoints and under sensory shuffle.

\FloatBarrier

\end{document}